\documentclass{article}
\usepackage{spconf,amsmath,epsfig,enumitem,tabto}
\usepackage{url}
\usepackage{booktabs}
\usepackage{xspace}
\usepackage{breakurl}
\usepackage[breaklinks]{hyperref}
\usepackage{subcaption}

\setlist{nosep}

\title{ESTIMATING DISPLACED POPULATIONS FROM OVERHEAD}

\name{\em{Armin Hadzic$^{1,2}$, Gordon Christie$^1$, Jeffrey Freeman$^1$, Amber Dismer$^3$, Stevan Bullard$^4$} \\ \em{Ashley Greiner$^3$, Nathan Jacobs$^2$, Ryan Mukherjee$^1$}}

\address{{\centering $^1$Johns Hopkins University Applied Physics Laboratory \quad $^2$University of Kentucky}\\{\centering $^3$Centers for Disease Control and Prevention \quad $^4$Agency for Toxic Substances and Disease Registry}}

\begin{document}

\newcommand{\reducecaptionspace}{\vspace{-0.3cm}\xspace}
\newcommand{\reducepostfigspace}{\vspace{-0.3cm}\xspace}
\newcommand{\reducespace}{\vspace{-0.3cm}\xspace}

\newcommand{\osmonly}{\textsc{osm-only}\xspace}

\maketitle

\begin{abstract}

We introduce a deep learning approach to perform fine-grained population estimation for displacement camps using high-resolution overhead imagery. We train and evaluate our approach on drone imagery cross-referenced with population data for refugee camps in Cox's Bazar, Bangladesh in 2018 and 2019. Our proposed approach achieves 7.02\% mean absolute percent error on sequestered camp imagery. We believe our experiments with real-world displacement camp data constitute an important step towards the development of tools that enable the humanitarian community to effectively and rapidly respond to the global displacement crisis.

\end{abstract}


\section{Introduction}
\label{sec:intro}

According to the United Nations High Commissioner for Refugees (UNHCR), there are currently 70.8 million people forcibly displaced worldwide~\cite{UNHCRstats}, which is the largest human displacement crisis in history. Many displaced persons find themselves living in camps where health and other basic human services are constrained and the threat of violence a persistent concern. The humanitarian community has mobilized significant resources to address the displacement crisis. However, responses are inherently reactionary and there is a strong desire for tools that enable a rapid and accurate assessment of populations in need during times of crisis. Deep learning with overhead imagery offers a practical solution to improve response. In this paper, we detail an approach for estimating camp population using overhead imagery, which can often be obtained rapidly and throughout an event. Population estimates provide a good starting point because they can feed into many different types of analysis, such as determining whether sufficient water, sanitation, and hygiene (WASH) facilities are present in a camp.

\begin{figure}[htb]
\begin{minipage}[b]{1.0\linewidth}
  \centering
  \centerline{\epsfig{figure=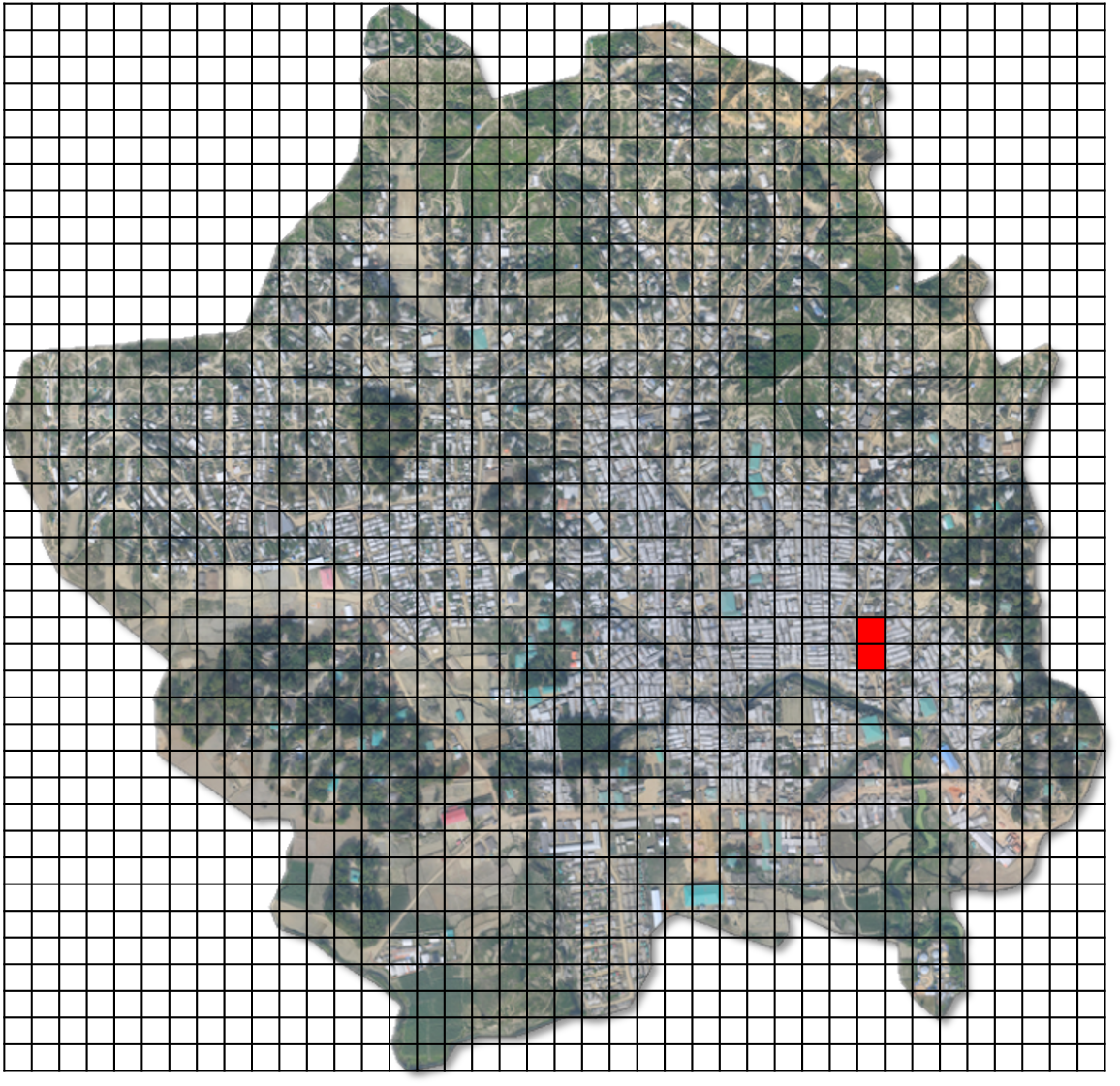,width=6.5cm}}
\end{minipage}
\hfill
\begin{minipage}[b]{1.0\linewidth}
  \centering
  \centerline{\epsfig{figure=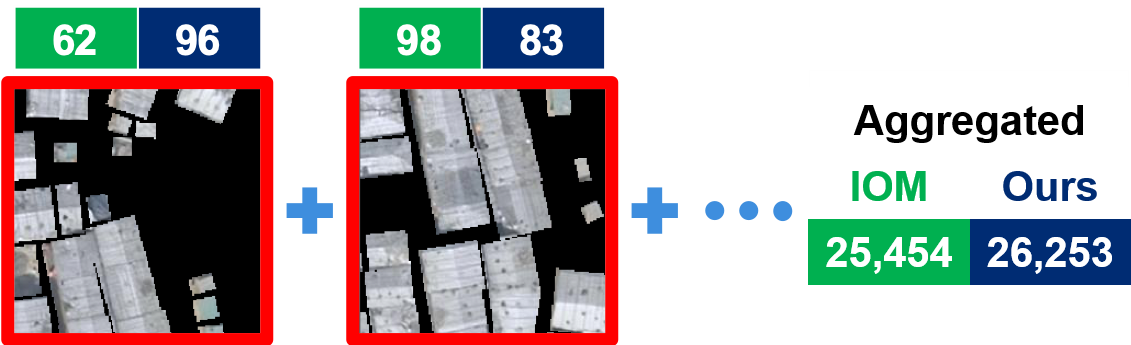,width=6.5cm}}
\end{minipage}
\caption{Generating test images (top) using equally spaced and sized image chips. Population predictions (blue) and IOM-reported population labels (green) are shown per chip. The predicted population of the entire camp (blue, bottom right) is the sum of all of the image chip predictions.}
\reducecaptionspace
\reducecaptionspace
\label{fig:TestingProcess}
\end{figure}

While population estimation using overhead imagery is not novel, high-resolution (i.e., 30cm ground sample distance and better) estimates have only recently become possible due to improved sensor capabilities. Given the increasing availability of such imagery, along with the recent advancements in computer vision and machine learning, we believe the time is ripe to start training and deploying high-resolution models for the humanitarian community. Most previous work estimate population at a
substantially coarser scale, leveraging approaches unlikely to transfer well to high-resolution camp imagery. \cite{robinson} disaggregate US census tract data into fine-grain cells using weighted combinations of intersecting census blocks, similar to our disaggregation approach. However,~\cite{robinson} also discretize population counts into 17 bins to perform classification and operate on coarse 15m Landsat imagery. \cite{ermonMapping} perform population estimation using imagery of India,
including rural areas that may be more visually similar to humanitarian camps, however they also use coarse resolution Landsat and Sentinel data and their approach struggles with fine-grain village-level ($\le$ 20.25km\textsuperscript{2} area) population estimation. \cite{jacobs2018population} introduce techniques for performing pixel-level population estimation with 3m resolution Planet imagery. However, these techniques are only applied to well-organized US cities and it is unclear how well
they might handle less-organized camp settings where the density and appearance of structures are drastically different. 

Related approaches using sub-meter resolution imagery include \cite{fmow2018} detecting impoverished settlements and \cite{kemperDarfur} counting dwellings in Darfur camps, both using 30-50cm DigitalGlobe imagery but neither estimating population. On the other hand, \cite{galeon2008estimation} use structure areas extracted from Quickbird imagery to predict population, but their method is mostly manual and lacks large-scale evaluation.

To our knowledge, our approach is the first to perform learning-based population estimation using sub-meter overhead imagery (10cm GSD). We have publicly released our code to train and test population estimation models, as well as generate a dataset based on open source imagery and population data\footnote{\scriptsize\url{https://github.com/JHUAPL/EstimatingDisplacedPopulations}}. Together, we believe these contributions constitute an important step towards the development of tools that enable the humanitarian community to effectively and rapidly respond to the global displacement crisis.

\begin{figure}[htb]
\begin{minipage}[b]{1.0\linewidth}
  \centering
  \centerline{\epsfig{figure=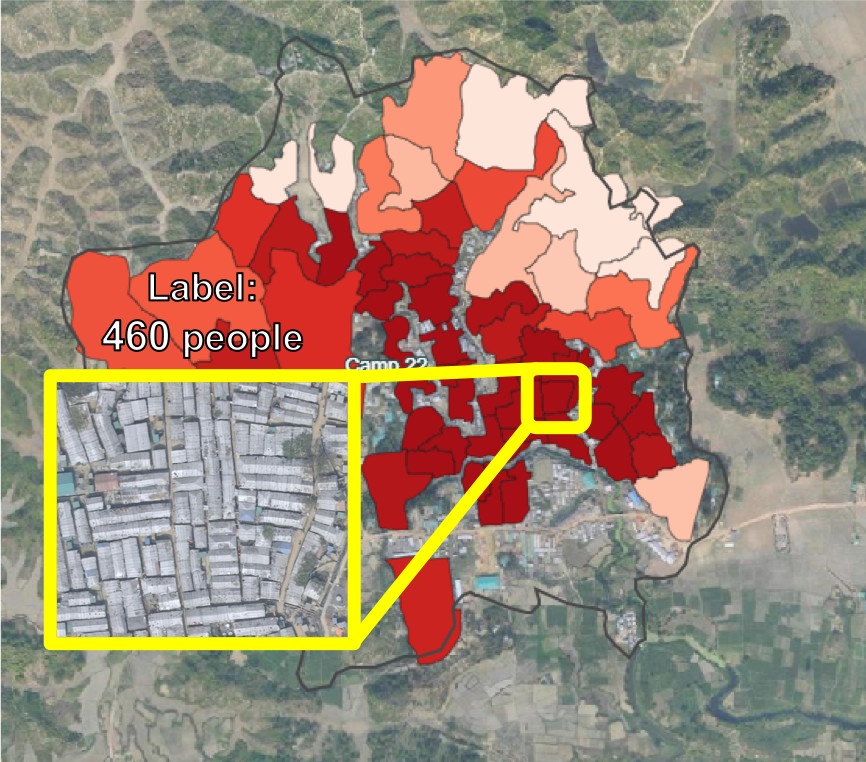,width=6.5cm}}
\end{minipage}
\caption{Majhee block (population polygon) labels from Camp 22 in Cox's Bazar, Bangladesh. Combining these labels produces an IOM-reported population label for an entire camp.}
\reducecaptionspace
\reducecaptionspace
\label{fig:majheeBlk}
\end{figure}

\section{Problem Statement}
\label{sec:probStatement}

Population data is traditionally provided in the form of census tracts, which vary in size and cover large spatial areas (Figure ~\ref{fig:majheeBlk}). However, fine-grain population mapping is desirable and beneficial for humanitarian efforts such as camp management. Pixel-level population labels are generally unavailable due to security concerns and annotation costs. We address this problem by using a simple area-based tract disaggregation method. Then to reduce the cost of conducting camp censuses altogether, we train a model using aerial imagery to directly predict camp population at high spatial resolution.

\section{Dataset}
\label{sec:dataset}

Our dataset is comprised of (1) overhead drone imagery, (2) population polygons (majhee blocks), and (3) OpenStreetMap (OSM) structure segmentation masks, all from refugee camps in Cox's Bazar, Bangladesh. Overhead images were tiled into square chips and paired with labels extracted from the population polygons. Population polygons correspond to majhee block data taken from routine International Organization for Migration (IOM) Bangladesh: Needs and Population Monitoring (NPM) site assessments. The data corresponding to 10\% of the 34 camps were sequestered to the test split.

\subsection{Overhead Imagery}
\label{ssec:overheadImgr}

Overhead imagery for our dataset was also sourced from NPM site assessments~\cite{droneImagery}. The imagery is comprised of georeferenced 10cm overhead drone images. Each of the 34 camps in the Cox's Bazar region have up to nine overhead images, each averaging a 2.7km\textsuperscript{2} area, and totaling 294 images. The nine possible images per camp correspond to different site assessments performed during different months and seasons across two years.

\begin{figure}[htb]
	\begin{subfigure}{0.49\columnwidth}
		\centering
		\includegraphics[trim=5px 95px 10px 30px, clip, width=\columnwidth]{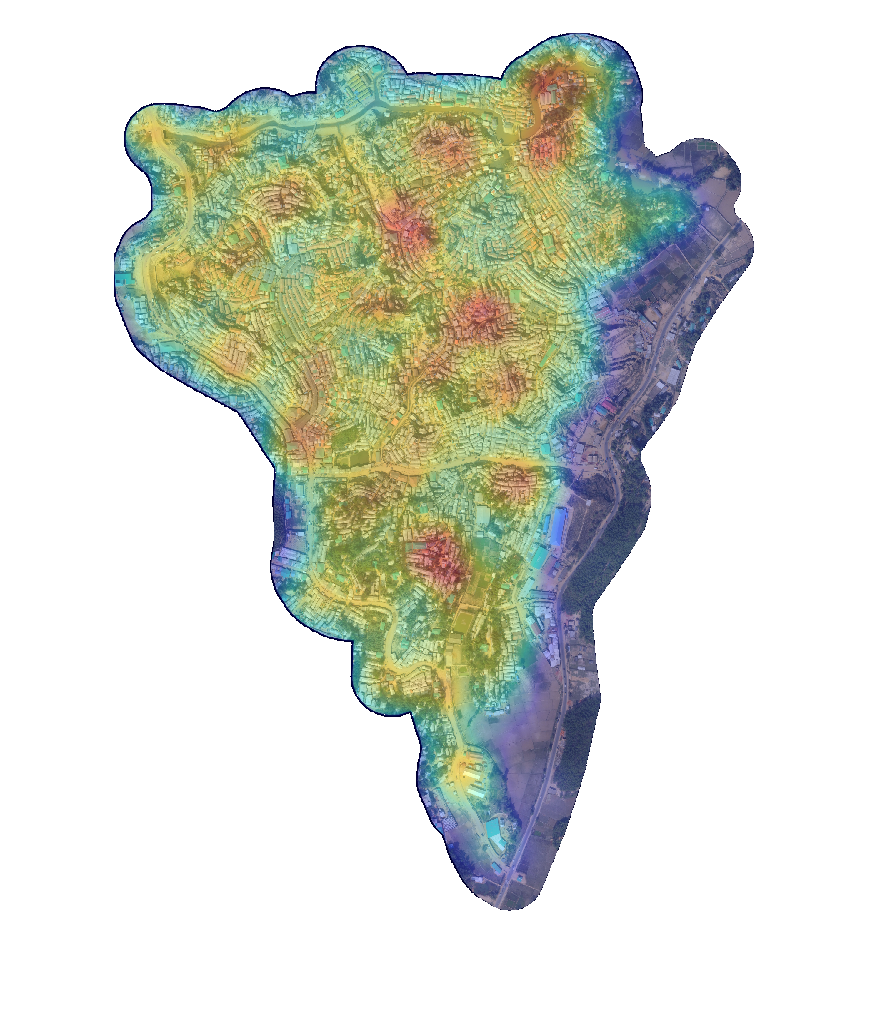}
		\caption{Our Prediction}
	\end{subfigure}
	\hspace*{\fill} % separation between the subfigures
	\begin{subfigure}{0.49\columnwidth}
		\centering
		\includegraphics[trim=5px 95px 10px 30px, clip, width=\columnwidth]{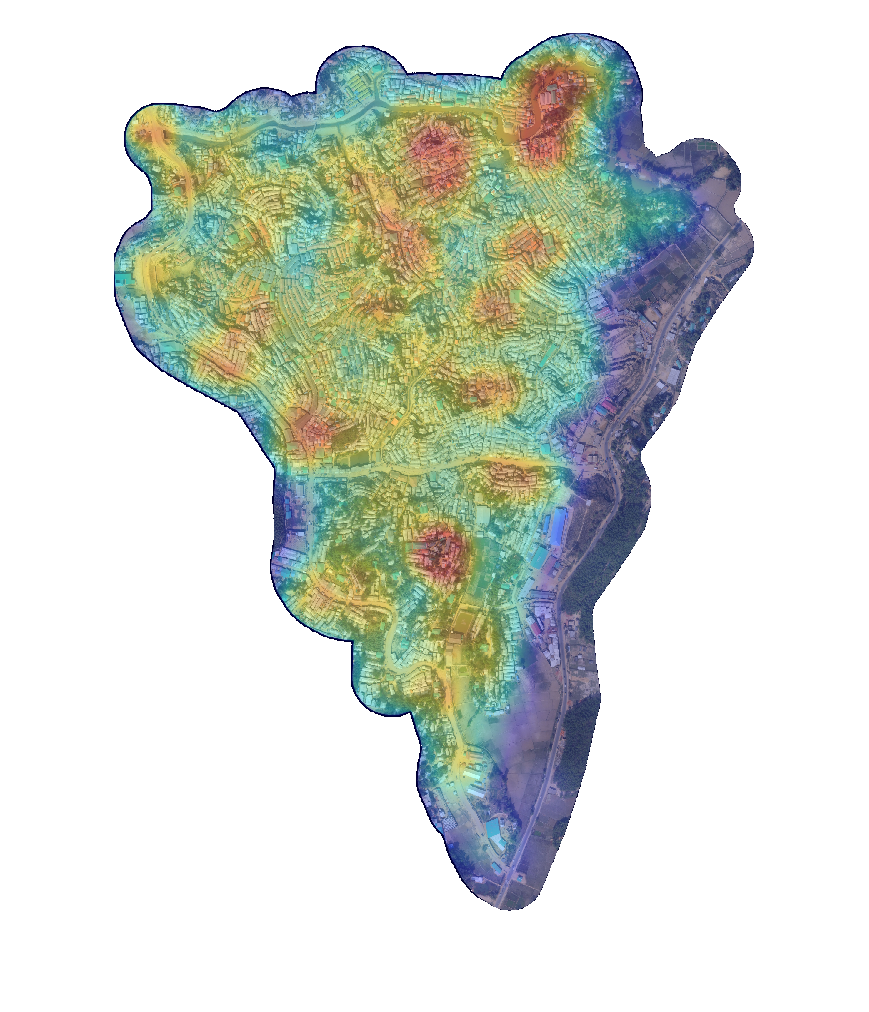}
		\caption{IOM Estimate}
	\end{subfigure}
	\caption{Qualitative comparison of model predictions and IOM estimates from March 2019 for Camp 11 in Cox's Bazar, Bangladesh. Red indicates relatively high population density, while blue indicates relatively low population density.} \label{fig:qualitative_comparison}
    \reducecaptionspace
\end{figure}

For each image in the training split, 200 square bounding boxes were extracted, stored as image chips, and used to train a convolutional neural network (CNN). Bounding boxes were both randomly positioned and sized, where a box's edge length $l$ is randomly chosen from the set $l \in \{224,  320,  480,  640,  720,  1024\}$ without exceeding the source image size. Randomly selecting bounding box locations and sizes provides several
benefits, such as increased quantity of training data and improved generalizability by training on imagery with slightly varying effective spatial resolutions and population densities. Input image chips were then resized, not cropped, from their original bounding box size down to 224x224 before being passed through the model.

Conversely, test split image chips are all of size 224x224 and were sampled in a sliding-window fashion to ensure overhead test images were fully covered with no overlap. As Figure~\ref{fig:TestingProcess} illustrates, the sliding-window chip generation approach allows a model to make a prediction on each image chip such that the sum of all predictions reflects the total displacement camp population estimate.

OSM's human-annotated structure masks were used to focus the model's feature extraction by removing non-relevant image pixels and help reduce scene appearance bias for when the model is applied to new scenes. We accomplished this by using OSMNX~\cite{OsmnxMaskGen} to retrieve and register structure annotations for each image chip (Figure~\ref{fig:TestingProcess}). Note that this registration process does result in some misalignment. Test chips containing structures with 0 corresponding population were omitted during test time as the buildings were outside Majhee block boundaries.

\subsection{Population Polygons}
\label{ssec:popPoly}

Population estimates for refugee camps in Cox's Bazar, Bangladesh are currently performed using majhee blocks, named after the majhee community leaders responsible for portions of each camp. Majhee blocks are represented as irregular polygons in the NPM shapefile dataset available on the Humanitarian Data Exchange~\cite{droneImagery}, as shown in Figure~\ref{fig:majheeBlk}. It is worth noting that manual population estimation is a challenging and time-consuming task. The majhee block system is not perfect due to the existence of some gaps in camp coverage and unquantified errors in population count. 

Nonetheless, image chip population labels ($y_{GT}$) are calculated using the weighted sum of majhee block polygons that overlap with a chip's bounding box, as described by Equation~\ref{eq:gtLbl} and similar to the methodology used by~\cite{robinson}. The square bounding box ($b_{i,j,l}$) is defined by its bottom-left corner pixel coordinate ($i$,$j$), and by length $l$. The spatial area of a bounding box or polygon is represented by $A_p$ and $c(z_k)$ is the population of a given polygon. Additionally, $z_k$ is the $k$th polygon from the set of polygons $n$ in the shapefile that most closely matches the date and time of when the overhead image was captured, as follows: 
\begin{gather}
    y_{GT}(b_{i,j,l})=\sum_{k=0}^{n} c(z_k) \dfrac{ A_p(z_k \cap b_{i,j,l})}  {A_p(b_{i,j,l})}. \label{eq:gtLbl}
\end{gather}

\section{Method}
\label{sec:method}

We perform displacement camp population estimation using a convolutional neural network (CNN). Our model approximates the population density $\hat{f}$, which can be represented as $\hat{f}(m(I_{i,j,l}) \land I_{i,j,l}, GSD;\Theta) \approx f(I_{i,j,l})$, where $I_{i,j,l}$ is the input image chip, $m(I_{i,j,l}) \land I_{i,j,l}$ is the image chip overlaid by the mask ($m$) chip, $GSD$ is the ground sample distance of the image chip, $A_c$ is the spatial area depicted in the chip, $\Theta$ is the set of model parameters, and $f$ is the known chip population density. The population density of the camp depicted in the chip is then translated into a chip scale population prediction, $\hat{P}(I_{i,j,l})$, using the spatial area of the chip, $\hat{P}(I_{i,j,l}) = \hat{f}A_c \approx P(I_{i,j,l})$. During inference all chips are uniformly sized ($l=224$ for $I_{i,j,l}$) and do not overlap, so the chips corresponding to a single full overhead image can be aggregated to obtain the total predicted camp population.

\section{Experimental Results}
\label{sec:results}

Quantitative performance is evaluated on entire camps since chip-scale population labels are unavailable. However, we perform a chip-level qualitative evaluation to assess our model's fine-grain prediction performance.

As a baseline, which we call \osmonly, we used a linear Huber regressor that predicts the population of an image chip  using the structure area ($A_s$) of the corresponding chip as an input. Similar to the CNN, the baseline performs fine-grained  predictions on chips, sums the predictions, and yields the full camp population estimation. Let $n$ be the number of structures ($s_r$) in a given structure mask chip. The total structure area ($A_t$) is calculated using the structure segmentation mask chip ($m(I_{i,j,l})$) of the corresponding image chip ($I_{i,j,l}$), as follows: 
\begin{equation}
A_t(m(I_{i,j,l})) = \sum_{r=0}^{n} A_s(s_r). \label{eq:structRatio}
\end{equation}

\noindent\textbf{Metrics:} Similar to~\cite{robinson}, we use Mean Absolute Error (MAE) to reflect the total number of individuals unaccounted for, and Mean Absolute Percent Error (MAPE) to generally reflect model performance.

\noindent\textbf{Implementation Details:} The proposed CNN model was built from a ResNet50 architecture pre-trained on ImageNet and modified for regression. The image chip's GSD was concatenated with the bottleneck features of the network attached to the regression head. Ground truth was represented as density (i.e., population normalized by area) instead of raw population to enable more robust training and testing of models with imagery of varying GSD and chip sizes. Training was performed with the Adam optimizer and a default initialized triangular cyclic learning Rate~\cite{smith2017cyclical}. Huber loss was used and found to work better than mean squared error or log-cosh loss. Additionally, the following image augmentations~\cite{albumentations} were applied to improve model generalization: vertical flip, random $90\deg$ rotation, CLAHE, random brightness, random gamma, hue saturation, and random contrast.

\noindent\textbf{Quantitative Evaluation:} The quantitative results are 3,704 MAE (10.00\% MAPE) and 3,341 MAE (7.02\% MAPE) for \osmonly and our approach, respectively. These results demonstrate that the CNN's image-based features are better-suited for predictive displacement camp populations over structure area alone.

\begin{figure}[htb]
\begin{minipage}[b]{1.0\linewidth}
  \centering
  \centerline{\includegraphics[trim=5px 5px 40px 30px, clip, width=\textwidth]{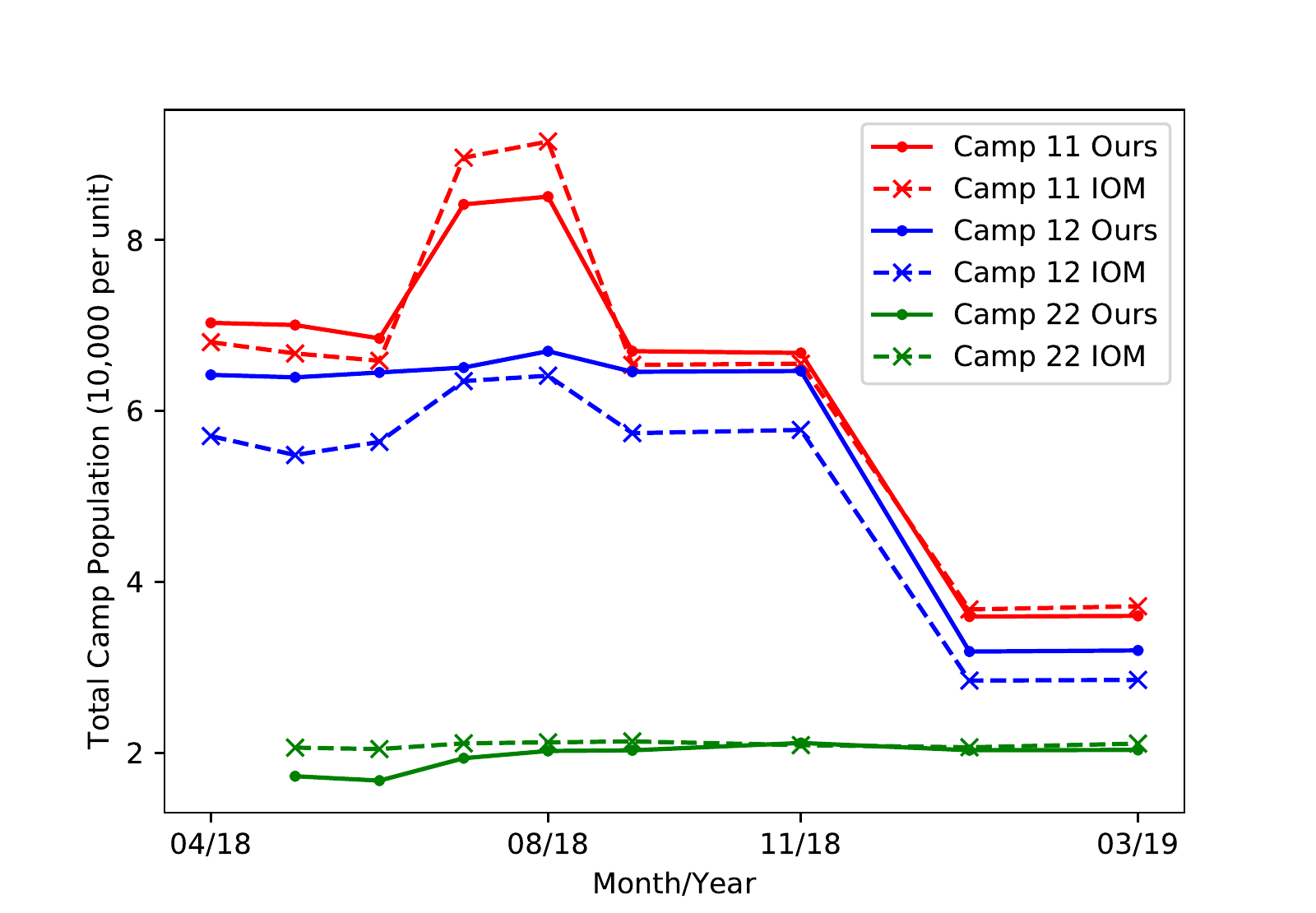}} % left, lower, right, upper
\end{minipage}
\caption{CNN model (solid line) performance on the three heldout camps (11, 12, and 22) compared to IOM-reported population (dashed).}
\reducecaptionspace
\reducecaptionspace
\label{fig:timeseriesQual}
\end{figure}

\noindent\textbf{Qualitative Evaluation:} Qualitative evaluation of our model is shown in Figure~\ref{fig:qualitative_comparison}. Overall, our model's predicted population distribution closely follows the IOM-reported counts. Our model tends to struggle with chips that are densely populated (dark red regions) as opposed to sparsely-populated regions (shades of blue). For fair comparison between the reported and predicted population, areas that do not overlap with majhee blocks are not shown. While our approach is capable of performing predictions in these areas, accuracy cannot be evaluated in said areas. Additionally, Figure~\ref{fig:timeseriesQual} shows the CNN total camp population performance compared with IOM-reported population data over time\footnote{Camp 22 did not have recorded data for 04/18. Data for months 10/18, 12/18, and 02/19 was incorporated into the following month's data.}. Note that imagery and population counts were recorded on a monthly basis for 8 months, and our method was able to predict an aggregated camp population that followed the contour of the reported population for all three camps. Additionally, our model did not strictly overpredict or underpredict for two of the three camps, suggesting our model is not unidirectionally biased.

\reducespace

\section{Conclusions}
\label{sec:conclusions}

\reducespace

We introduce a new dataset for developing fine-grained population estimation models using high-resolution drone imagery of refugee camps in Cox's Bazar, Bangladesh. Using this dataset, we developed a novel approach capable of achieving 7.02\% mean absolute population estimation error on a sequestered test set. The success of our approach demonstrates that structure masks can be used to encourage networks to learn correlations between building image features and population estimates. Future work could incorporate: methods for structure segmentation~\cite{ye2019building} and more rigorous disaggregation techniques~\cite{jacobs2018population}. Nevertheless, we believe this approach and dataset constitute an important step towards the development of tools that enable the humanitarian community to effectively and rapidly respond to the global displacement crisis.

\reducespace

\bibliographystyle{IEEEbib}
\bibliography{strings,refs}

\end{document}